%% file: main.tex
\documentclass{article}

\PassOptionsToPackage{authoryear, round, compress}{natbib}

\usepackage[preprint]{neurips_2025}
\usepackage{tcolorbox}
\tcbuselibrary{breakable}

\usepackage[utf8]{inputenc} 
\usepackage[T1]{fontenc}    
\usepackage{hyperref}       
\usepackage{url}            
\usepackage{booktabs}       
\usepackage{amsfonts}       
\usepackage{nicefrac}       
\usepackage{microtype}      
\usepackage{xcolor}         

\usepackage{geometry}
\usepackage{array}
\usepackage{amsmath}
\usepackage{amsthm}
\theoremstyle{definition}
\newtheorem{definition}{Definition}[section]
\usepackage[capitalize,noabbrev]{cleveref}

\usepackage{graphicx}
\usepackage{subcaption}

\title{Darwin Mobile Agent: A Roadmap for Self-Evolution}

\author{
  \textbf{Daniel Beechey\textsuperscript{1,*},} 
  \textbf{Derek Yuen\textsuperscript{1,*},} 
  \textbf{Jianheng Liu\textsuperscript{1,*},} 
  \textbf{Dezhao Luo\textsuperscript{1},} \\
  \textbf{Tiantian He\textsuperscript{2},} 
  \textbf{Weilin Luo\textsuperscript{1},} 
  \textbf{Jun Wang\textsuperscript{2},} 
  \textbf{Kun Shao\textsuperscript{1,$\dagger$}} \\
  \textsuperscript{1}Huawei Noah's Ark Lab, \textsuperscript{2}University College London
}

\begin{document}

\renewcommand{\thefootnote}{\fnsymbol{footnote}}
\footnotetext{\textsuperscript{*}Equal Contribution, \textsuperscript{$\dagger$}Corresponding author: shaokun2@huawei.com}
\renewcommand{\thefootnote}{\arabic{footnote}}

\maketitle

\begin{abstract}
The goal of artificial intelligence is to create agents capable of general, adaptive behaviour in open-ended environments. Guided by the ``Bitter Lesson'', we argue that the most effective path toward this goal is to systematically remove human priors and allow intelligence to naturally emerge through interaction with a ``Big World'' that is orders of magnitude more complex than the agent itself. We propose the mobile Graphical User Interface (GUI) as a practical proxy for such a world and introduce \emph{Darwin Mobile Agent}, an open-source infrastructure designed as a foundation for autonomous reinforcement learning in this domain. This framework addresses the data-collection bottleneck in real-world mobile interactions by using an asynchronous agent-environment loop across parallel cloud-phone instances. We further propose a conceptual roadmap to systematically remove human priors from three fundamental pillars of a self-evolving agent: task curricula, outcome verification, and memory management. We validate that the Darwin infrastructure provides the stability and scalability required for the first stage of this roadmap: policy optimisation in the GUI domain. This work establishes the practical and theoretical foundation necessary to move toward truly autonomous, self-evolving GUI agents.
\end{abstract}

\input{sections/section_introduction}

\input{sections/section_background}

\input{sections/section_env}

\input{sections/section_framework}

\input{sections/section_results}

\input{sections/section_conclusion}

\newpage
\bibliography{references}
\bibliographystyle{plainnat}

\appendix

\input{sections/appendix}

\end{document}

%% file: sections/section_introduction.tex
\section{Introduction}
\label{sec:introduction}

The long-standing goal of artificial intelligence is to create agents capable of general, adaptive behaviour in open-ended environments. History suggests the most effective path toward this goal is to develop methods that scale with computation rather than relying on specific human knowledge---an observation known as the ``Bitter Lesson'' \citep{sutton2019bitter}. This implies that we must systematically remove human priors from the learning process. Equally important is the nature of the problem itself. Inspired by bounded rationality \citep{simon1955behavioral, simon1956rational} and the ``Big World'' hypothesis \citep{javed2024big,lewandowski2025world}, we argue that general intelligence emerges as a necessary adaptation when a bounded agent interacts with a world orders of magnitude more complex than itself. In such a regime, the agent cannot perceive the global state or learn a perfect policy for every situation; it must approximate and generalise from limited experience.

Our long-term objective is to apply these principles by designing a \emph{self-evolving agent} capable of continuous, autonomous capability expansion solely through interaction. In this report, we address the first two fundamental infrastructure challenges:
\begin{enumerate}
    \item \textbf{The World:} We require an experimentally tractable domain that is sufficiently open-ended to qualify as a ``Big World'', mirroring the structure and complexity of the real world.
    \item \textbf{The Agent Learning Framework:} We require a learning and interaction architecture capable of sustaining an autonomous learning loop.
\end{enumerate}

\paragraph{The world.} We propose the Mobile Graphical User Interface (GUI) as a practical ``Big World'' for this investigation. The modern mobile ecosystem acts as a digital proxy for the real world: it is partially observable, highly non-stationary, and offers a nearly infinite number of composable tasks \citep{klissarov2025discovering}. Unlike static benchmarks (e.g. Atari), mobile apps evolve independently of the agent, requiring continual adaptation. Yet, unlike physical robotics, the domain remains purely digital, allowing us to leverage the massive semantic knowledge of Large Language Models (LLMs) without the constraints of physical embodiment.

We introduce a Gymnasium-based GUI environment \citep{towers2025gymnasium} that relies on cloud-based devices to mitigate the instability we observed in ADB-based emulators \citep{toyama2021androidenv,rawles2024androidworld}. In contrast to existing frameworks, our environment avoids reliance on underlying XML phone states to ensure scalability. Furthermore, our interaction model is \emph{asynchronous}, enabling the parallelism required to address data-collection bottlenecks in large-scale reinforcement learning.

\paragraph{The agent learning framework.} In accordance with the ``Bitter Lesson'', we contend that reinforcement learning (RL) is essential for developing self-evolving agents capable of learning through trial-and-error interaction. We propose that sustaining general, open-ended learning within this framework necessitates at least three additional components:
\begin{itemize}
    \item \textbf{Task Curriculum:} A mechanism to propose a stream of problems at the frontier of the agent's capabilities, providing a steady gradient of complexity.
    \item \textbf{Verification:} A mechanism to assess outcomes and generate the reward signals that drive learning.
    \item \textbf{Agent State:} A mechanism to persist and organise knowledge, ensuring the agent retains context across a non-stationary world (e.g. memory).
\end{itemize}

\begin{figure}[t]
    \centering
    \includegraphics[width=1.0\linewidth]{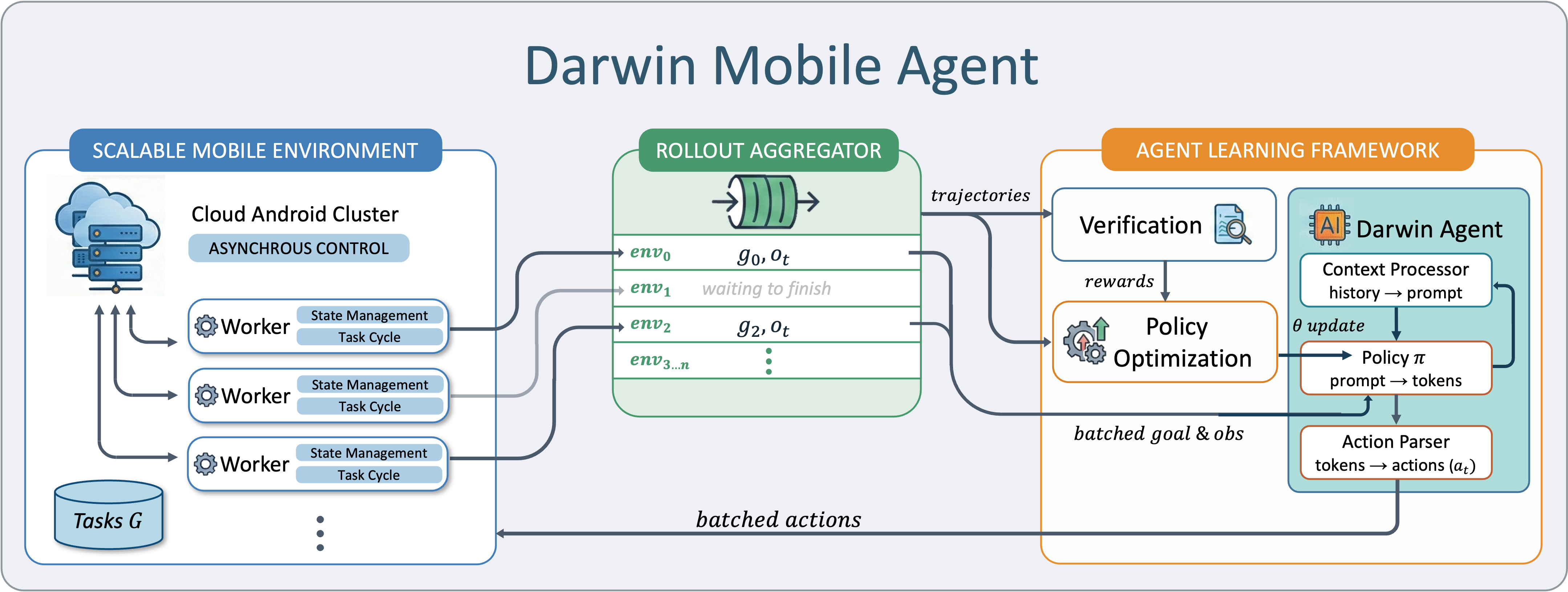}
    \caption{\textbf{Overview of the Darwin Mobile Agent Framework.} The system employs an asynchronous \emph{Rollout Aggregator} to bridge the gap between slow, parallelised mobile environments and high-throughput agent inference. Completed trajectories are passed to a verification module to generate rewards for continuous policy optimisation.}
    \label{fig:diagram}
\end{figure}

We refer to these components as the functional \emph{anatomy} of the self-evolving agent. To support this structure, we introduce \emph{Darwin Mobile Agent}%
\footnote{\href{https://github.com/ai-agents-2030/darwin-mobile-agent}{github.com/ai-agents-2030/darwin-mobile-agent}}
(\cref{fig:diagram}), a multi-turn learning framework built upon GiGPO~\citep{feng2025group} and VERL~\citep{sheng2025hybridflow}. The framework employs high-level abstractions for \emph{agents} and \emph{workflows} to decouple internal state (e.g. memory) from interaction logic (e.g. curriculum), providing the modularity required for autonomously improving the anatomy's specific components. To our knowledge, this represents the first open-source end-to-end system for training mobile GUI agents via reinforcement learning.

We demonstrate the system's effectiveness by fine-tuning a UI-TARS agent using reinforcement learning on tasks from Spa-Bench \citep{chenspa}. These experiments verify that our framework and environment provide the stable, scalable infrastructure required to improve capabilities in complex GUI environments. We also investigate how critic bias impacts training stability and demonstrate that our asynchronous design improves throughput without compromising learning progress.

The remainder of this paper is organised as follows: \cref{sec:background} provides background; \cref{sec:environment} details the scalable environment; \cref{sec:learning_framework} presents the agent learning framework and our proposed anatomy; \cref{sec:results} discusses experimental results; and \cref{sec:conclusion} concludes.

%% file: sections/section_background.tex
\section{Background}
\label{sec:background}

Inspired by work on general reinforcement learning \citep{hutter2005universal,hutter2024introduction} and recent work moving beyond the Markov property \citep{dong2022simple,abel2023definition,abel2024three,elelimy2025rethinking}, we adopt a general agent-environment formulation. To support our goal of self-evolving agents, we purposefully make minimal assumptions about underlying dynamics, accommodating non-stationary, non-Markovian, and partially observable settings.

\subsection{The Agent-Environment Interface}
We first define the interaction boundary between the agent and the environment \citep{dong2022simple}.

\begin{definition}
    The \textbf{interface} is a pair of finite sets $(\mathcal{A}, \mathcal{O})$, such that $|\mathcal{A}| \geq 2$ and $|\mathcal{O}| \geq 2$.
\end{definition}

This interface creates a space of interaction sequences. We call any sequence up to a discrete decision point $t \in (0, 1, \dots)$ a history \citep{hutter2005universal,dong2022simple,abel2023definition}.

\begin{definition}
    A \textbf{history} $h_t \in \mathcal{H}$ is a sequence of observations and actions:
    \begin{equation}
        h_t = (o_0, a_0, o_1, a_1, \dots, o_t).
    \end{equation}
\end{definition}

We define the \emph{agent} and \emph{environment} as stochastic functions over these histories \citep{russell1994provably}.

\begin{definition}
    An \textbf{agent} is a function $\lambda: \mathcal{H} \to \Delta(\mathcal{A})$ mapping a history to a probability distribution over actions.
\end{definition}

\begin{definition}
    An \textbf{environment} is a function $e: \mathcal{H} \times \mathcal{A} \to \Delta(\mathcal{O})$ mapping a history and an action to a probability distribution over observations.
\end{definition}

At each decision point, the agent observes $o_t \sim e(h_{t-1}, a_{t-1})$ and selects an action $a_t \sim \lambda(h_t)$. Following the reward hypothesis \citep{sutton1998reinforcement,bowling2023settling}, we assume a scalar reward signal can capture all of the agent's goals and purposes.

\begin{definition}
The \textbf{reward function} $r: \mathcal{H} \times \mathcal{A} \to \mathbb{R}$ maps a history and action to a scalar feedback signal.
\end{definition}

We make no assumptions about the source of the reward; it may be external (e.g. from the environment) or internal (e.g. from the agent). The agent aims to learn a policy that maximises a statistic of this reward. Here, we focus on the expected discounted sum of rewards (expected return):
\begin{equation}
    \mathbb{E} \left[ \sum_{t=0}^{T} \gamma^t r_t \right],
\end{equation}
where $\gamma$ is a discount factor.

This formulation captures goal-conditioned objectives. By embedding a semantic goal $g$ (e.g. natural language instructions) in the observation $o$, the reward function---operating on the history of these observations---implicitly becomes a function of the goal.

\subsection{Large Language Models as Agents}
We implement the abstract agent $\lambda$ using Large Language Models (LLMs). This requires distinguishing between high-level environment actions $a \in \mathcal{A}$ and the model's low-level generation process. The LLM uses a fixed vocabulary of tokens $\mathcal{V}$. We denote the set of all finite sequences of these tokens as $\mathcal{V}^*$. We parameterise the agent's policy $\pi_\theta$ using a neural network with weights $\theta$.

At step $t$, the model generates a token sequence $y_t = (y_t^0, y_t^1, \dots, y_t^K)$ autoregressively:
\begin{equation}
    \pi_\theta(y_t | h_t) = \prod_{k=0}^{K} \pi_\theta(y_t^k | \psi(h_t), y_t^{<k}),
\end{equation}
where $y_t^{<k}$ are the tokens generated so far in the current step. We define the \emph{agent state} as $s_t = \psi(h_t)$, where a context processing function $\psi: \mathcal{H} \to \mathcal{V}^*$ maps interaction history to a finite context window. This state effectively acts as the agent's memory, transforming raw history into a discrete prompt (e.g. via truncation or summarisation).

The generated sequence $y_t$ is not the environment action itself. It may contain reasoning or formatting. We therefore define a deterministic parser $\xi: \mathcal{V}^* \to \mathcal{A}$ to extract the executable command:
\begin{equation}
    a_t = \xi(y_t).
\end{equation}
This explicitly separates the agent's internal reasoning ($y_t$) from its external influence ($a_t$).

\subsection{Policy Optimisation}

In our experiments, we learn parameters $\theta$ to maximise expected return using policy gradient methods \citep{sutton1999policy}. The gradient of the objective $J(\theta)$ is defined as:
\begin{equation}
    \nabla_\theta J(\theta) = \mathbb{E}_{\pi_\theta} \left[ \sum_{t=0}^{T} \sum_{k=0}^{|y_t|-1} \nabla_\theta \log \pi_\theta(y_t^k | s_t, y_t^{<k}) A^{\pi_\theta}(s_t, y_t^{<k}) \right],
\end{equation}
where $A^{\pi_\theta}(s_t, y_t^{<k}) = Q^{\pi_\theta}(s_t, y_t^{<k}) - V^{\pi_\theta}(s_t, y_t^{<k})$ is the advantage function, which quantifies the relative value of a specific token compared to the baseline state-value.

We employ Proximal Policy Optimisation (PPO) to optimise this objective \citep{schulman2017proximal}. PPO maximises a surrogate objective via importance sampling, reweighting transitions based on the divergence between the current policy and the policy used during data collection. We maximise the clipped surrogate loss:
\begin{equation}
    \mathcal{L}(\theta) = \mathbb{E} \left[ \sum_{t=0}^{T} \sum_{k=0}^{|y_t|-1} \min \left( \rho_t^k(\theta) \hat{A}_{t,k}, \text{clip}(\rho_t^k(\theta), 1-\epsilon, 1+\epsilon) \hat{A}_{t,k} \right) \right].
\end{equation}
Here, $\rho_t^k(\theta)$ is the importance sampling ratio between policies $\pi_\theta$ and $\pi_{\text{rollout}}$:
\begin{equation}
    \rho_t^k(\theta) = \frac{\pi_\theta(y_t^k | s_t, y_t^{<k})}{\pi_{\text{rollout}}(y_t^k | s_t, y_t^{<k})}.
\end{equation}
This clipping mechanism enforces a trust region, preventing large, destabilising policy updates.

We estimate the token-level advantage $\hat{A}_{t,k}$ using Generalised Advantage Estimation (GAE) \citep{schulman2015high} and a learned value function $V_\phi(s_t, y_t^{<k})$. This function approximates the expected return conditioned on the agent's state and partial token sequence. The advantage $\hat{A}_{t,k}$ is the exponentially weighted sum of temporal difference (TD) errors $\delta_{t,k}$, computed at the token level:
\begin{equation}
    \delta_{t,k} = \begin{cases}
    \gamma V_\phi(s_t, y_t^{<k+1}) - V_\phi(s_t, y_t^{<k}) & \text{if } k < |y_t|-1, \\
    r_t + \gamma V_\phi(s_{t+1}) - V_\phi(s_t, y_t^{<k}) & \text{if } k = |y_t|-1.
    \end{cases}
\end{equation}
Values for intermediate tokens are bootstrapped from the subsequent token; the final token incorporates the reward $r_t$ and is bootstrapped from the next decision step. Value function parameters $\phi$ are optimised by minimising the mean squared error between the predicted values and the empirical returns:
\begin{equation}
    \mathcal{L}^{VF}(\phi) = \mathbb{E} \left[ \sum_{t=0}^{T} \sum_{k=0}^{|y_t|-1} (V_\phi(s_t, y_t^{<k}) - \hat{R}_{t,k})^2 \right].
\end{equation}

Alternatively, we could use approaches like GRPO to estimate the trajectory-level advantage and assign it equally to each token \citep{shao2024deepseekmath}. While effective for bandit-like problems \citep{guo2025deepseek}, failing to distinguish individual token and action contributions is inevitably detrimental in prolonged, multi-turn interactions.

%% file: sections/section_env.tex
\section{A Scalable Asynchronous Mobile Environment}
\label{sec:environment}

In this section, we instantiate the agent-environment interface as a Gymnasium environment \citep{towers2025gymnasium} for mobile GUIs. We detail our cloud-based infrastructure and asynchronous execution model, selected for stability and scalability.

The environment's general interface is independent of proprietary internals, mirroring human interaction. Observations $o_t$ consist of raw pixel screenshots and task instructions $g$. The action space $\mathcal{A}$ comprises primitive coordinate-based operations (e.g. $\text{click}$, $\text{drag}$). Our current iteration also includes specific high-level operations, such as $\text{open-app}$ and $\text{type}$, to improve training efficiency. A complete list of supported actions is provided in Appendix~\ref{appendix:action_space}. This specification is exposed via the standard Gymnasium interface.

We implement this interface on a fleet of cloud-hosted Android instances, controlled via the Android Debug Bridge (ADB). We prefer this infrastructure over local emulation \citep{toyama2021androidenv,rawles2024androidworld} or physical farms \citep{li2025mobileuse} because it avoids hardware constraints on app coverage, enables flexible device scaling, and we find that dedicated, predictable cloud resources offer greater stability.

In this high-fidelity setting where smartphone agents must operate at human-scale latency, environment interaction, not model inference, is the primary training bottleneck. We use vectorised interaction to achieve the necessary throughput. However, mobile step durations vary significantly due to network latency, device processing, and disconnections. Standard synchronous vectorisation forces the entire batch to wait for the slowest device, stalling the learning process and wasting GPU resources. To prevent this, we propose a scalable asynchronous architecture.

\subsection{Cloud-Native Asynchronous Architecture}

To realise this asynchronous design, we decouple environment workers from policy inference. Each worker manages a dedicated cloud device in an isolated process. Upon completing a step, the worker immediately pushes the result to a central inference queue and continues execution.

This architecture eliminates the rigid phase separation of synchronous RL, where the GPU remains idle while waiting for rollouts. Instead, it enables continuous pipelining: the learner consumes batches from the queue as soon as the minimum batch size is met, allowing concurrent GPU computation and environment execution. Fast devices continue generating experience even when others are temporarily blocked. This design enables massive scaling, effectively rendering interaction latency negligible through parallelism. Consequently, the number of workers is decoupled from the inference batch size. The system operates on small, responsive batches from whichever workers are ready, rather than waiting for a massive vector to synchronise. In the mobile domain---where latency is high and variable---this minimises wall-clock wait times and improves hardware utilisation.

\subsection{Persistent State Management}

A central challenge is managing device state across episodes. Standard reinforcement learning resets the system to a pristine state at the start of each episode. On mobile devices, a complete reset (e.g. reflashing the system image) takes several minutes, making it infeasible even for asynchronous architectures at scale. To maintain throughput, our environment performs minimal cleanup by returning to the home screen after each episode. This introduces \emph{persistence}: changes to the file system or clipboard may persist into the next episode.

Crucially, the mobile domain involves state \emph{external} to the device, such as social media posts, which cannot be locally reverted. A ``perfectly clean state'' is unrealistic. We treat this non-stationarity as a feature: by exposing the agent to persistent ``noise'' and irreversible state, we encourage the development of robust policies capable of operating in real-world ecosystems. However, to prevent state saturation from rendering a finite number of tasks trivial, our initial experiments apply periodic hard resets via the cloud provider's APIs.

\subsection{Task Definition and Lifecycle}
\label{subsec:task_lifecycle}
We define a task as a tuple containing the natural-language instruction $g$ and metadata, such as the target application. This lightweight representation supports diverse tasks by making minimal assumptions about the underlying system (see Appendix \ref{appendix:implement_details} for examples).

Realistic tasks often have preconditions or leave side effects. Because we preserve state, we cannot rely on automatic resets to handle these. Instead, we introduce a \emph{Task Lifecycle Protocol}. Configured by optional task parameters, this decomposes interaction into three phases, making state management the agent's responsibility:
\begin{enumerate}
    \item \textbf{Setup Phase:} The agent establishes preconditions (e.g. ensuring a file exists).
    \item \textbf{Execution Phase:} The agent fulfills the primary instruction $g$.
    \item \textbf{Teardown Phase:} The agent restores the environment to a neutral state (e.g. deleting a created file).
\end{enumerate}

Transitions are success-gated: the environment advances only when the agent completes the current phase. This internalisation enables agents to learn complex, long-horizon workflows without artificial external resets. A new task is sampled when the agent exhausts its step budget or signals completion. We distinguish these terminations from infrastructure failures, which trigger non-punitive truncations to decouple agent performance from transient device instability.

%% file: sections/section_framework.tex
\section{The Agent Learning Framework}
\label{sec:learning_framework}

In \cref{sec:introduction}, we argued that a self-evolving reinforcement learning agent needs three core components: a task curriculum, outcome verification, and persistent state. Standard reinforcement learning usually treats these as fixed properties of the environment. To build an agent that improves continuously, we need a fundamental shift in architecture: we must move these mechanisms out of the environment and define them instead as dynamic properties of the agent itself.

This shift requires a flexible architecture that can handle \emph{any} definition of agent state or interaction logic. To address this, we introduce Darwin Mobile Agent, a \emph{multi-turn learning framework} built on GiGPO~\citep{feng2025group} and VERL~\citep{sheng2025hybridflow}. It uses two primary abstractions to model agent--environment interaction: \emph{Agents}, which handle the internal decision-making, and \emph{Workflows}, which manage the interaction logic. By separating the loop's coordination from the environment, this design allows us to insert custom logic---such as dynamic task sampling or internal reward generation---directly into the interaction process.

This modularity supports an iterative research strategy where external implementations, such as curated task sets and LLM judges, are systematically replaced with internalised, learned equivalents. In this section, we first detail our architectural abstractions and optimisation loop. We then define our functional \emph{anatomy}---comprising curriculum, verification, and state---and discuss the progressively removing human priors to enable fully autonomous learning.

\subsection{Environment Interaction as Agents and Workflows}
\label{subsec:agents_and_workflows}

The \texttt{Agent} interface transforms observations into executable actions. It manages the decision process through two main functions. First, context processing ($\psi: \mathcal{H} \to \mathcal{S}$) turns the interaction history into a model-compatible state representation $s_t$ (i.e. the prompt). This abstraction allows us to use diverse, modular memory types (e.g. a sliding window or an evolving summary) without changing the policy implementation. The model then processes this state to generate a token sequence $y_t$. Second, action projection ($\xi: \mathcal{V}^* \to \mathcal{A}$) maps this output to the environment's action space. Together, these components encapsulate the specific prompting strategy and output formatting (e.g. JSON schemas or coordinate-based DSLs), ensuring the environment interface or learning algorithm remains agnostic to the underlying model backend.

The \texttt{Workflow} interface manages the agent--environment loop, acting as a general layer sitting above the agent. Its primary purpose is to support diverse interaction patterns by separating the control logic from the environment definition. For example, a workflow can coordinate complex action-selection behaviours, such as hierarchical planning or multi-agent collaboration, treating them as internal subroutines that eventually produce a final action $a_t$. Crucially, this generality allows us to implement mechanisms for dynamic task proposal and internal outcome verification, moving these responsibilities from the environment into the agent's interaction loop.

\subsection{The Rollout and Optimisation Loop}
\label{subsec:rollout_loop}

We now consider the rollout loop, which coordinates the continuous interaction between the agent-workflow system and a distributed set of environment instances to generate experience for learning. At each step, the workflow processes a batch of observations to sample actions. These actions are executed in parallel, and the outcomes are processed to generate a reward signal. The resulting atomic transition tuples $(o_t, a_t, r_t)$ are then pushed asynchronously to a centralised replay buffer. Unlike synchronous rollout loops that wait for all environments to terminate \citep{feng2025group}, this design treats instances independently; as soon as a task finishes, a new one is sampled immediately to ensure zero idle time.

When the centralised buffer accumulates sufficient data, we trigger an update by retrieving only fully completed trajectories, keeping active episodes in the buffer for subsequent steps. We adopt this strategy to avoid the value-estimation bias introduced by truncating trajectories and bootstrapping from the critic in the middle of a task. We find that relying on actual final rewards is necessary to provide a more stable signal than these intermediate approximations. While the retrieved data is off-policy by the time we use it, our experiments show that the stability gained from ground-truth rewards outweighs the benefits of remaining strictly on-policy.

During the optimisation phase, we temporarily reconstruct the retrieved step data into temporally ordered trajectories. This allows us to perform accurate credit assignment by using GAE to estimate advantages at the granularity of individual tokens over the entire interaction horizon. This approach improves upon prior methods \citep{feng2025group} that neglect sequential dependencies by estimating advantages for steps in isolation, while avoiding the limitations of rigid frameworks \citep{sheng2025hybridflow} that treat the full trajectory as a single continuous datapoint. By establishing the atomic transition as the fundamental unit of data, we separate the optimisation mechanics from the interaction structure, ensuring the learning process scales to arbitrary task-horizons unconstrained by the model's context window.

\subsection{Anatomy of the Self-Evolving Agent}
\label{subsec:agent_anatomy}

The proposed architecture for our self-evolving agent comprises three core components: a task curriculum, outcome verification, and persistent state. While the bitter lesson suggests that agents should not rely on human priors, existing research often builds these mechanisms using knowledge external to the agent. Whether they rely on hand-designed task sets \citep{rawles2023androidinthewild}, infrastructure-specific verification \citep{rawles2024androidworld}, or manually engineered context management \citep{ye2025mobile}, these dependencies prevent the emergence of a truly autonomous system.

Our research goal is to systematically replace these external priors with learned equivalents. We propose a three-stage trajectory in which responsibility for each component shifts from humans to autonomous external models and, finally, to the agent itself. While the second stage removes the human from the loop, the agent remains inherently capped by the static capabilities of the external model. True self-evolution requires internalising these processes so that they evolve alongside the agent. In this subsection, we detail the role of each component, its current implementation in Darwin Mobile Agent, and a path toward removing all reliance on external knowledge.

\paragraph{Task Curriculum.}

In an open-ended setting, the objective is not to master a fixed set of problems, but to sustain continuous learning. This objective requires a task curriculum that manages (1) what tasks are presented and (2) when they are introduced. Its goal is to maintain a steady learning gradient exactly at the frontier of the agent's capabilities. This gradient is fragile: if a task is too simple, the agent has insufficient signal to refine its policy; if it is too difficult, the learning process stagnates due to a lack of learning signal. Our experiments show that policy refinement is most effective within a narrow range of problems characterised by a low, but non-zero, initial success rate.

We propose a three-stage trajectory for internalising the curriculum and removing its reliance on external knowledge. In the first stage, human designers observe agent performance to identify and propose learnable problems. The second stage uses a more capable external model to automate task selection by evaluating task learnability based on the agent's interaction history. Finally, the agent (meta-)learns to manage the curriculum internally, removing the fixed performance ceiling imposed by external supervision.

In the current version of Darwin Mobile Agent, we adopt the first stage of this trajectory. We manually select a diverse set of tasks from SPA-Bench \citep{chenspa} where our base models exhibit non-zero but low success rates. This ensures the learning gradient necessary to verify that our environment and optimisation loop effectively improve the agent's capabilities.

\paragraph{Outcome Verification.}

For a reinforcement learning problem to be well-posed, the system must be able to verify whether an agent has achieved its goal. Outcome verification provides the fundamental reward signal required to drive the learning loop, establishing a ground truth independent of secondary challenges such as shaped or sparse rewards. Unlike standard reinforcement learning environments, which often rely on programmatic state verification, such functions do not exist in domains like the mobile GUI that aim to emulate real-world complexity. Verifying success requires a semantic understanding of whether the agent’s actions fulfilled the intent of the instruction. Because this cannot be hard-coded for every possible goal, verification becomes a reasoning problem of comparable complexity to achieving the goal itself.

We apply the three-stage trajectory to outcome verification by systematically shifting the responsibility for verification from external sources to the agent. In the first stage, verification relies on human-designed programmatic functions that do not generalise across environments because they require direct access to internal system states \citep[e.g.][]{rawles2024androidworld}. The second stage uses an external model as an automated judge to enable semantic reasoning over open-ended tasks, though this remains constrained by the model's capabilities. Finally, the agent learns to verify its own outcomes, allowing its understanding of success to evolve alongside its improving capabilities.

In our current implementation of Darwin Mobile Agent, we adopt the second stage of this trajectory using an LLM-based judge. The primary challenge in this stage is determining the optimal inputs for the judge, specifically regarding the prompt structure and the extent of the interaction history provided. By validating various configurations against human-labelled ground truth from the SPA-Bench dataset (see Appendix \ref{sec:judge_configs}), we ensure that our initial judge offers a stable, accurate signal for the specific task distribution used in our experiments, where programmatic verification is unavailable.

\paragraph{Agent State.}

In complex domains such as mobile GUIs, immediate observations are rarely sufficient for effective decision-making; actions often depend on information accumulated across previous interactions. The agent requires a context-processing function $\psi$ to persist and organise its raw interaction history $h_t$ into a finite agent state $s_t$. This state must manage a spectrum of information: from short-term \emph{episodic} memory tracking the immediate sequence of interactions, to the \emph{semantic} and \emph{procedural} knowledge that constitutes an agent's behavioural priors. This long-term knowledge, internalised over a lifetime of experience, encompasses diverse patterns, including general execution rules and app-specific functional details. Ultimately, the agent's state design determines how effectively the agent can compress its experience into a representation that informs its current decision.

We apply a similar three-stage research trajectory to the agent state. In the first stage, the context-processing function $\psi$ is human-designed, using fixed heuristics, such as sliding-window buffers or predefined schemas, to select which interactions remain in the prompt. The second stage introduces an external model to act as a memory manager; this LLM-driven process reviews interaction histories to distil episodic trajectories into concise summaries and extract higher-level knowledge, such as app-specific functional patterns. Finally, in the third stage, the agent learns to manage its internal state by autonomously performing this distillation and extraction. By learning to optimise its own representation of history, the agent maintains an efficient and informative state without the need for external supervision.

In our current implementation of Darwin Mobile Agent, we adopt the first stage of this trajectory using a configurable episodic buffer as the context-processing function $\psi$. This mechanism maintains a sliding window of the $N$ most recent interactions, with the depth of the action sequence and visual history governed by the model's specific context constraints. For models that generate semantically meaningful outputs, such as Qwen3-VL \citep{bai2025qwen3vltechnicalreport}, we record the history using these natural-language action summaries rather than raw coordinate data. This ensures the agent state remains informative even when visual observations are truncated due to context limits.

%% file: sections/section_results.tex
\section{Results}
\label{sec:results}

Our evaluation provides initial validation of the Darwin infrastructure's stability, scalability, and robustness. Rather than focusing on peak performance against established benchmarks, we aim to demonstrate that our asynchronous training pipeline provides a reliable environment for reinforcement learning on mobile devices.

Unless otherwise specified, all experiments use a \texttt{UI-TARS-7B} base model \citep{qin2025ui} trained on a subset of eight tasks derived from SPA-Bench \citep{chenspa} (full details in Appendix \ref{appendix:implement_details}). These tasks are selected based on moderate initial success rates to provide a meaningful learning signal. Training curves are smoothed using a sliding-window mean of the last 20 points. In our plots, bold lines represent these averages, while fainter lines show the original data. To validate the infrastructure, we conduct six targeted investigations:

\begin{enumerate}
    \item \textbf{Reward Signal Calibration:} We assess the alignment of our LLM-based judge with human evaluation.
    \item \textbf{Infrastructure Stability:} We verify the framework's ability to support stable policy improvement on a fixed task set.
    \item \textbf{Distributed Scalability:} We examine how the system handles increased interaction throughput by scaling the number of concurrent cloud-phone instances.
    \item \textbf{Learning Scalability:} We assess how the learning process scales as we increase the number of tasks in proportion to the available hardware.
    \item \textbf{Task Lifecycle Validation:} We evaluate the system's ability to manage the automated setup and teardown phases introduced in \cref{subsec:task_lifecycle}.
    \item \textbf{Learning Robustness:} We examine how critic bias and initialisation strategies impact learning stability and prevent policy collapse.
\end{enumerate}

\subsection{Validation of the LLM-Judge Reward Signal}
\label{sec:judge_validation}

Before evaluating the training framework, we must establish the reliability of the automated reward signal. We validated our LLM-based verification judge against a human-labelled ground truth of 296 trajectories from the SPA-Bench dataset (147 English; 149 Chinese). Human annotators provided binary success labels by reviewing the natural-language goal alongside the complete interaction history.

To identify the most reliable reward signal for our pipeline, we compared various judge prompts and history configurations as detailed in Appendix \ref{sec:judge_configs}. The optimal configuration, which uses the full visual and action history, achieves an F1 score of 0.91 (90\% accuracy) on English tasks and 0.85 (89\% accuracy) on Chinese tasks. These results suggest that the automated judge provides a stable and accurate signal for the specific task distribution used in our experiments, justifying its use as the supervisor for our reinforcement learning pipeline.

\subsection{Stability of Policy Optimisation}
\label{sec:stability_baseline}

We start by verifying that the core reinforcement learning loop, comprising asynchronous interaction, verification, and policy optimisation, is stable under standard conditions. To evaluate this, we use a controlled setting in which eight tasks are executed concurrently on eight cloud phones. This configuration allows us to establish a baseline for policy improvement before investigating more complex system scaling behaviours.

\begin{figure}[h]
    \centering
    \includegraphics[width=0.7\linewidth]{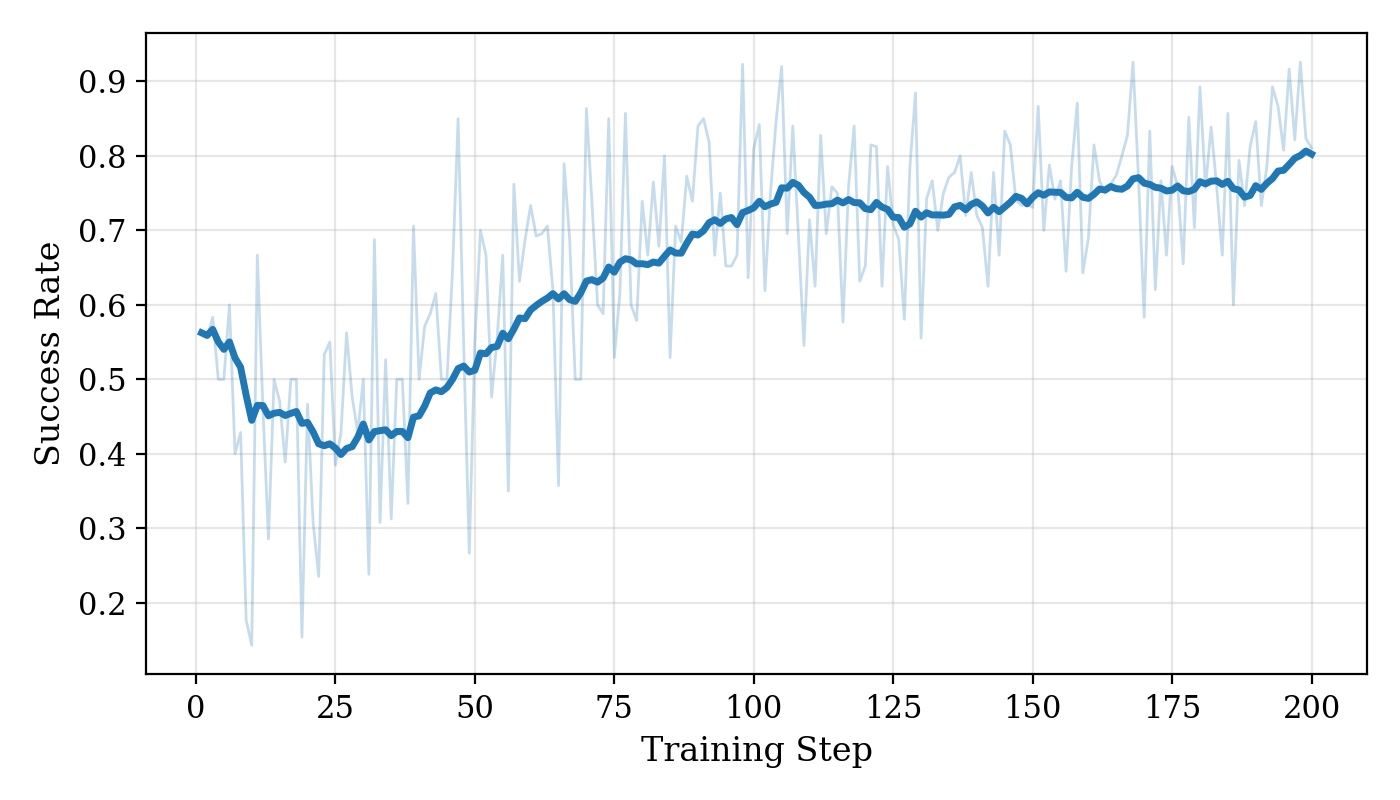}
    \caption{\textbf{Mean Success Rate on SPA-Bench.} The plot illustrates the average performance across eight tasks. Each training step represents 256 environment steps (8 phones $\times$ 32 steps per rollout), matching the maximum task horizon. The first 30 steps constitute a critic warm-up phase where the policy parameters are fixed.}
    \label{fig:stability_baseline}
\end{figure}

As shown in \cref{fig:stability_baseline}, the agent exhibits a clear upward trend in the mean success rate over the course of training. During the first 30 training steps, we implement a critic warm-up phase to initialise the value function. During this period, the actor parameters remain frozen; consequently, any fluctuations in the success rate are due to environmental and sampling variance and do not reflect policy updates. Following this warm-up, the agent demonstrates consistent improvement across the task set. These results suggest that the Darwin infrastructure can maintain a stable policy optimisation loop in the mobile GUI domain, providing a functional baseline for investigating the system's scalability and robustness.

\subsection{Distributed Scalability and System Throughput}
\label{sec:distributed_scaling}

We investigate the impact of hardware scaling on training throughput and learning stability by varying the number of concurrent cloud phones while maintaining a fixed task set and rollout size. This experiment aims to determine how the infrastructure handles increased parallelism in interactions and the resulting off-policy data drift. We evaluate three configurations: 8, 16, 32 concurrent phones, with the rollout size fixed at 256 steps across all settings.

\begin{figure}[t]
    \centering
    \begin{subfigure}[t]{0.48\linewidth}
        \centering
        \includegraphics[width=\linewidth]{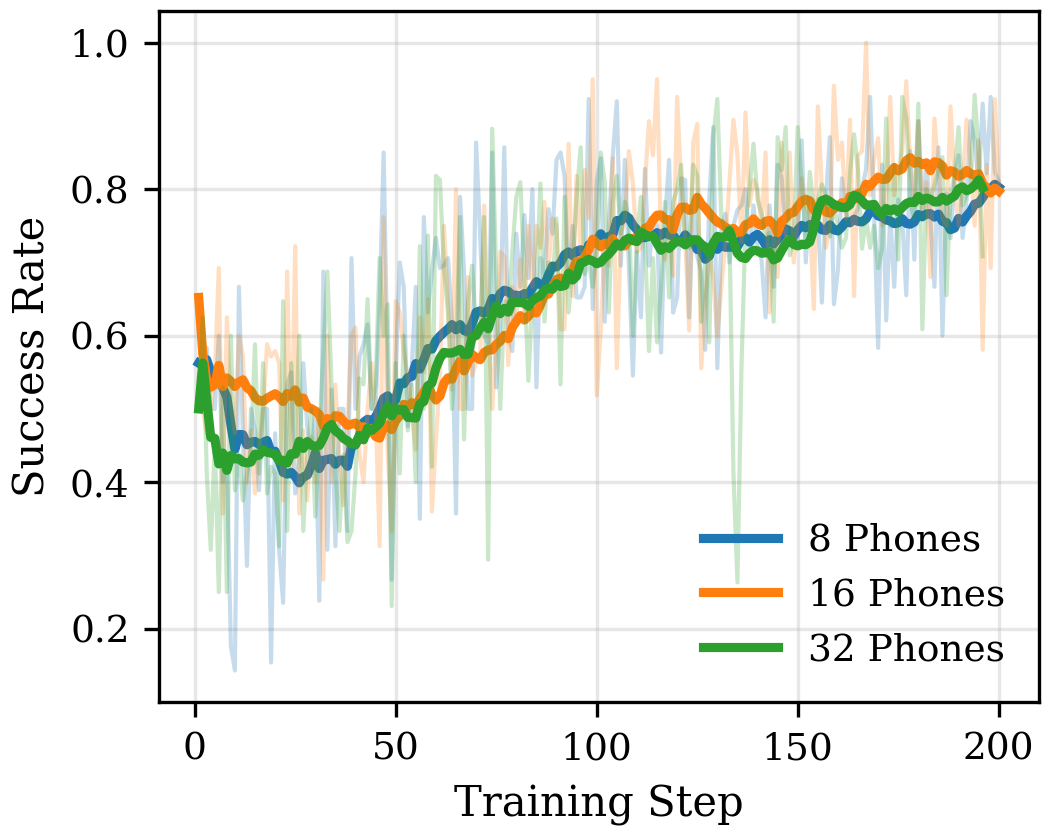}
        \caption{Mean success rate vs. training steps.}
        \label{fig:scaling_success}
    \end{subfigure}
    \hfill
    \begin{subfigure}[t]{0.48\linewidth}
        \centering
        \includegraphics[width=\linewidth]{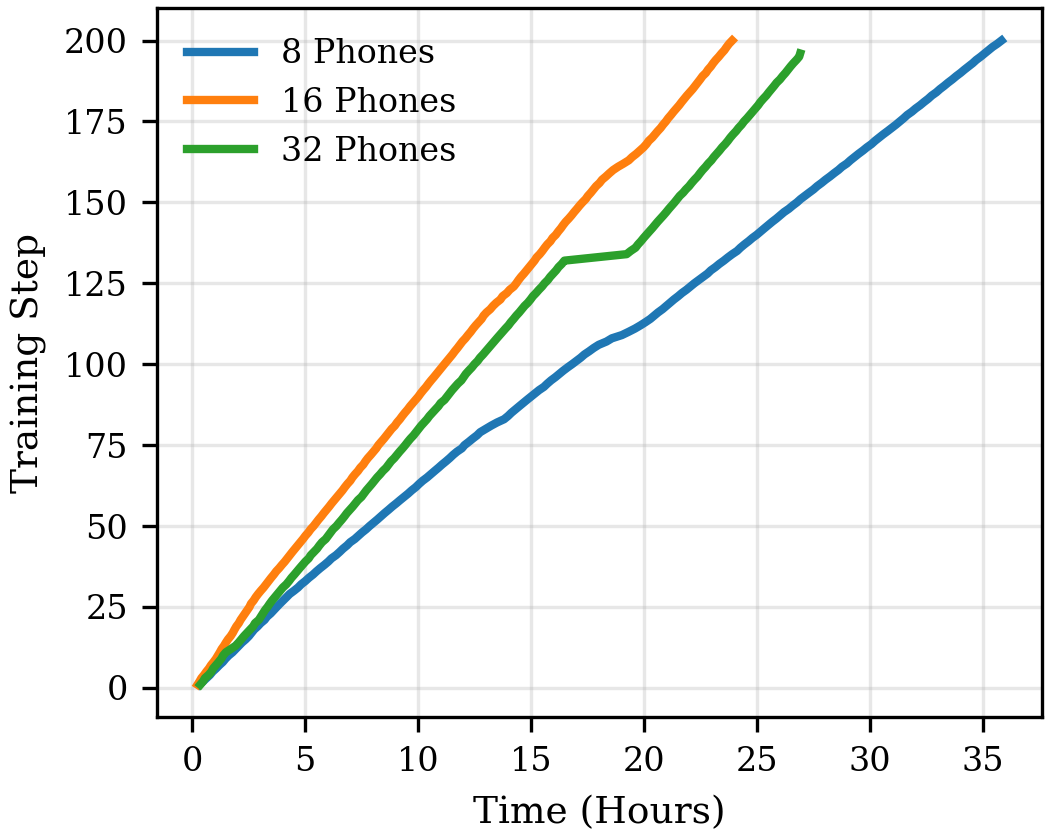}
        \caption{Training steps vs. wall-clock time.}
        \label{fig:scaling_time}
    \end{subfigure}
    \caption{\textbf{Distributed Scalability Evaluation.} (a) Scaling from 8 to 32 phones shows consistent convergence despite increased off-policy data. (b) Throughput improves from 8 to 16 phones but saturates at 32, as the system bottleneck shifts to model inference and batching overhead.}
    \label{fig:distributed_scaling}
\end{figure}

The asynchronous buffer accumulates experience until the 256-step threshold is reached; however, to avoid bootstrapping from a biased critic, the system waits for complete trajectories from terminated episodes. With higher device counts, this leads to significant over-collection per update. This excess experience is carried forward into subsequent training steps, increasing the proportion of off-policy data in the rollout buffer. As shown in \cref{fig:scaling_success}, this drift does not degrade learning effectiveness, as all configurations exhibit comparable convergence to the same mean success rate. Furthermore, the 32-phone configuration demonstrates resilience to hardware instability; despite a major network outage mid-training (evidenced by the plateau in \cref{fig:scaling_time}), the agent resumed learning without destabilisation once connectivity was restored.

\cref{fig:scaling_time} illustrates the corresponding impact on wall-clock throughput. We observe a reduction in training time when scaling from 8 to 16 phones. However, these gains saturate at 32 phones, as indicated by a training gradient similar to that in the 16-phone setting (excluding the device downtime). This suggests that at this scale, the system bottleneck shifts from environment interaction to the compute constraints of model inference, such as the increased batching overhead required to process more parallel streams. These results suggest that the Darwin infrastructure can robustly accelerate training through parallelism while remaining resilient to the transient failures inherent in real-world mobile environments.

\subsection{Task Scalability}
\label{sec:task_scaling}

To evaluate the infrastructure's capacity to handle increased environmental complexity, we investigate how the system scales when the number of tasks increases in proportion to available hardware. We compare a baseline configuration of 8 tasks executed on 16 concurrent phones against an expanded setting of 16 tasks on 32 phones. In the 16-task setting, we double both the total experience collected per training step and the PPO batch size to maintain a constant resource-to-task ratio across the two experiments.

\begin{figure}[h]
    \centering
    \includegraphics[width=0.7\linewidth]{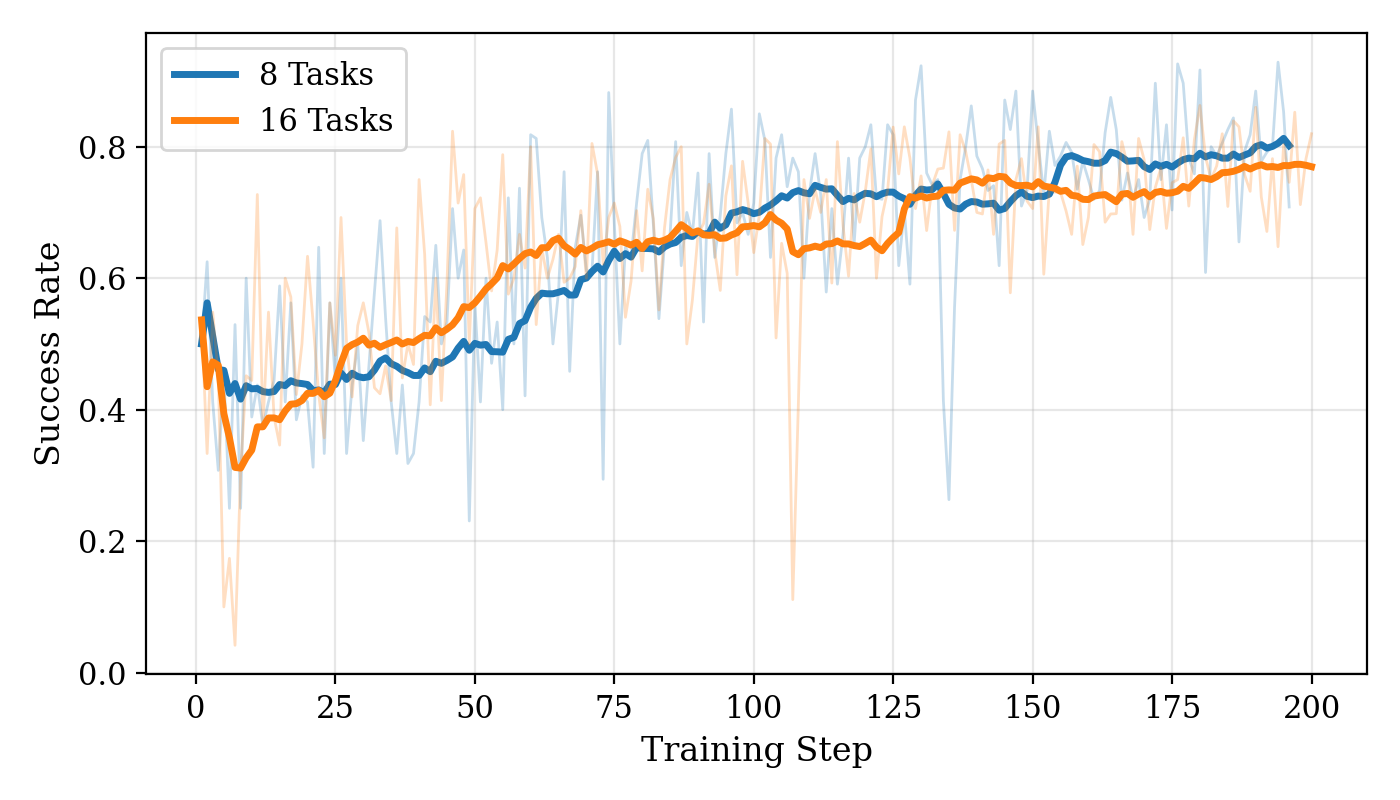}
    \caption{\textbf{Horizontal Scalability across Task Sets.} The plot compares the mean success rate when training on 8 tasks (16 phones) versus 16 tasks (32 phones). Despite the increased task diversity, both configurations exhibit similar convergence rates.}
    \label{fig:eval_num_tasks}
\end{figure}

As shown in \cref{fig:eval_num_tasks}, the learning curves for both the 8-task and 16-task configurations are similar. While each curve represents a mean over a different task distribution, they converge within the same number of training steps. These results suggest that the Darwin infrastructure supports horizontal scaling; the learning progression is preserved when hardware resources and optimisation hyperparameters are scaled proportionally with task diversity. This relationship indicates that the framework provides a predictable path for ``Big World'' training, where environmental complexity can be managed by incrementally adding hardware to maintain consistent learning performance.

\subsection{Task Lifecycle Validation}
\label{sec:lifecycle_validation}

To ensure the infrastructure can maintain environment consistency over extended training durations without manual intervention, we validate the automated task lifecycle protocol described in \cref{subsec:task_lifecycle}. This experiment evaluates the system and agent's ability to navigate the setup and teardown phases required for automated environment resets. We compare a baseline setting using standard environment resets against the ``cycle'' protocol. Both configurations employ 8 tasks executed on 16 concurrent cloud phones.

\begin{figure}[t]
    \centering
    \begin{subfigure}[t]{0.32\linewidth}
        \centering
        \includegraphics[width=\linewidth]{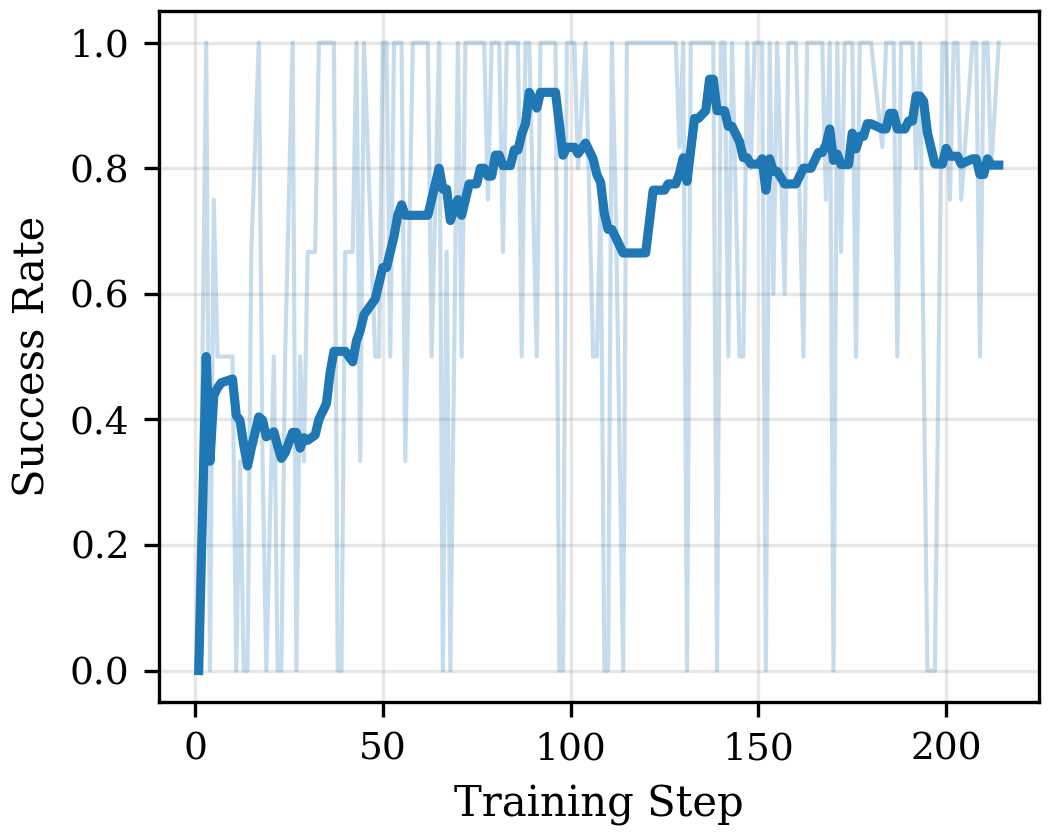}
        \caption{Setup Phase}
        \label{fig:lifecycle_setup}
    \end{subfigure}
    \hfill
    \begin{subfigure}[t]{0.32\linewidth}
        \centering
        \includegraphics[width=\linewidth]{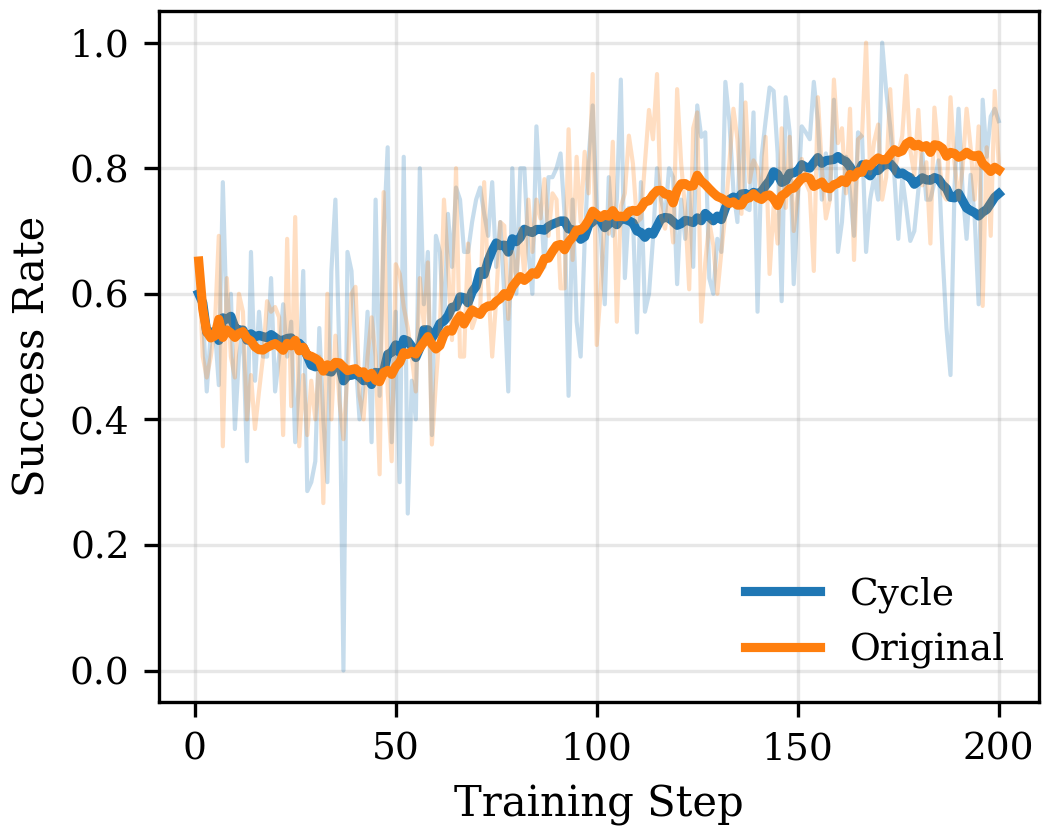}
        \caption{Task Execution}
        \label{fig:lifecycle_task}
    \end{subfigure}
    \hfill
    \begin{subfigure}[t]{0.32\linewidth}
        \centering
        \includegraphics[width=\linewidth]{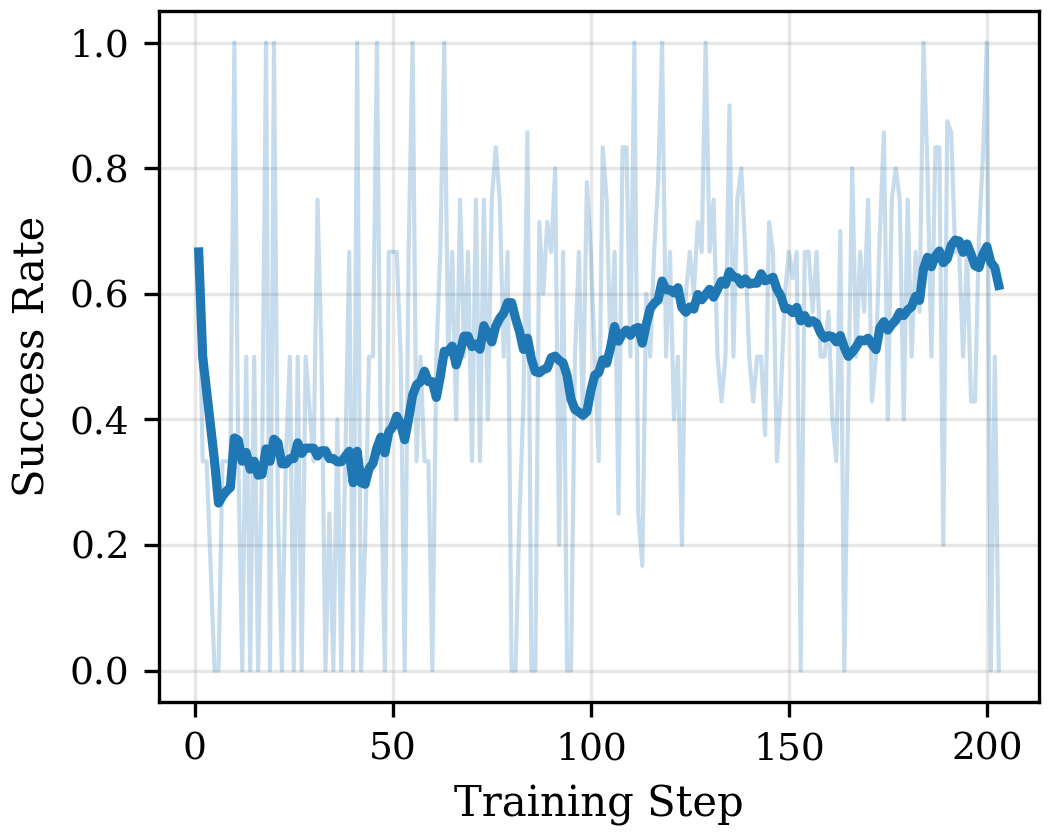}
        \caption{Teardown Phase}
        \label{fig:lifecycle_teardown}
    \end{subfigure}
    \caption{\textbf{Task Lifecycle Stability.} The panels illustrate the mean success rate across the three protocol phases. The ``cycle'' protocol successfully manages automated transitions; the agent exhibits a learning trend during the setup and teardown phases that mirrors the progression of the primary task.}
    \label{fig:task_lifecycle_results}
\end{figure}

The task set is heterogeneous: three tasks require no auxiliary phases, one includes a setup phase, and four include teardown phases. As shown in \cref{fig:task_lifecycle_results}, the inclusion of these explicit phases does not introduce instability or performance degradation. The mean success rate in the primary task execution phase (\cref{fig:lifecycle_task}) remains comparable between the two protocols, indicating that the agent acquires these auxiliary skills within the same total environment interaction budget without a trade-off in core task performance. Furthermore, the agent demonstrates a clear upward learning trend in both the setup and teardown phases (\cref{fig:lifecycle_setup,fig:lifecycle_teardown}). These results validate the task lifecycle protocol as a functional mechanism for autonomous, long-term training in the mobile GUI domain.

\subsection{Critic Bias and Bootstrapping Stability}
\label{sec:critic_robustness}

In multi-turn reinforcement learning for large language models, the critic is often the primary source of instability. Due to the combination of sparse rewards and long horizons, GAE-based advantage estimates rely heavily on bootstrapping from the critic to estimate state values. Consequently, policy improvement is highly sensitive to the accuracy of these value estimates. We investigate this sensitivity by evaluating the impact of critic initialisation and bootstrapping.

\begin{figure}[t]
    \centering
    \begin{subfigure}[t]{0.48\linewidth}
        \centering
        \includegraphics[width=\linewidth]{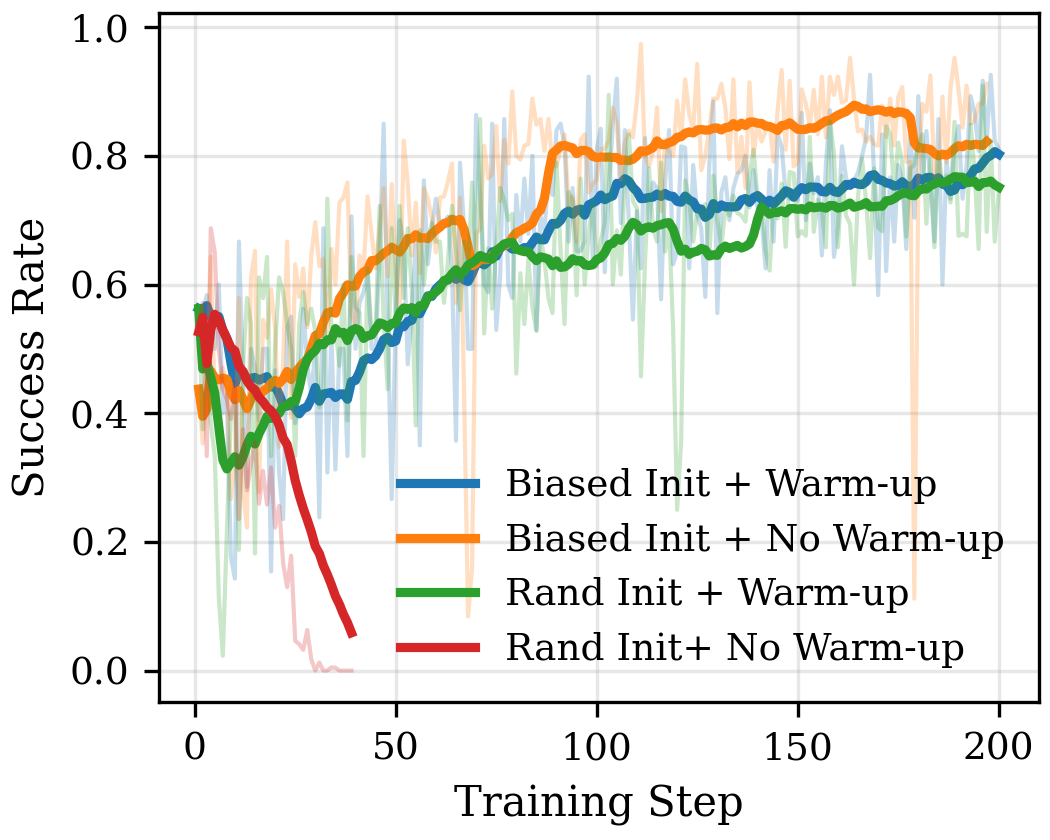}
        \caption{Initialisation and Warm-up.}
        \label{fig:critic_init}
    \end{subfigure}
    \hfill
    \begin{subfigure}[t]{0.48\linewidth}
        \centering
        \includegraphics[width=\linewidth]{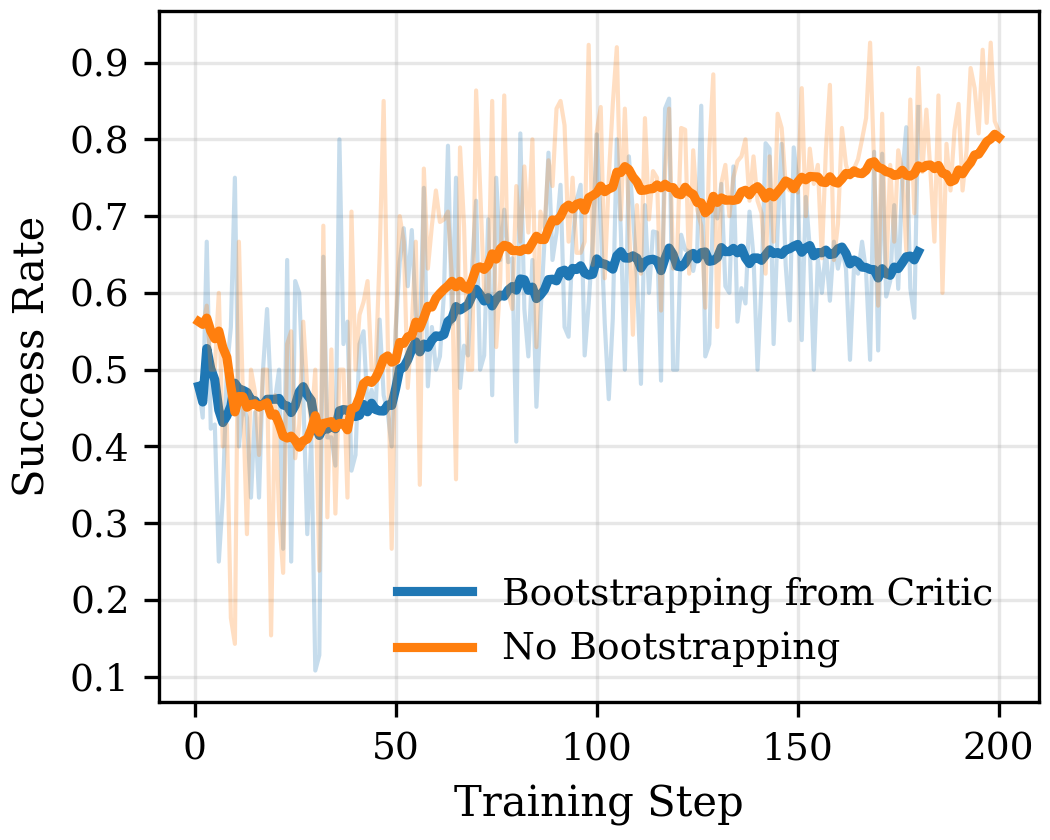}
        \caption{Truncation Handling.}
        \label{fig:bootstrap_comparison}
    \end{subfigure}
    \caption{\textbf{Critic Robustness Evaluation.} (a) Calibrated initialisation prevents policy collapse; stability is further supported by combining this initialisation with a warm-up phase. (b) Treating truncated trajectories as terminal by assuming the next state-value is zero ($V(s')=0$) outperforms bootstrapping from a learned critic.}
    \label{fig:robustness_results}
\end{figure}

As shown in \cref{fig:critic_init}, a randomly initialised critic without a warm-up phase leads to immediate policy collapse. Because returns in this domain are bounded between 0 and 1, random initialisation can produce values outside this range, generating incorrect advantage estimates that provide a destabilising signal to the policy. Consistent with previous findings \citep{yue2025vapo}, we find that a warm-up phase prevents this collapse by allowing the critic to reach a more accurate state before the policy is updated. Surprisingly, initialising the critic bias to $0.1$ ($\sigma=0$) provides a comparable safeguard. This calibrated initialisation alone prevents collapse and even enables faster learning by allowing for immediate policy updates. While these results suggest that calibrated initialisation can be sufficient, we recommend including a warm-up phase as a safeguard against an inaccurate initial bias.

This sensitivity to critic bias is further evidenced by the handling of trajectories truncated by external factors, such as system crashes, rather than reaching a natural conclusion. Although GAE theory supports bootstrapping from the critic to estimate the next state-value $s'$ at such points, \cref{fig:bootstrap_comparison} suggests that treating these truncations as terminal states with a future value of zero yields faster learning and higher final success rates than relying on potentially inaccurate critic estimates. These results suggest that in multi-turn LLM tasks, a predictable, pessimistic bias may be safer for estimating advantages and updating the policy than an unreliable bootstrap from the learned value function.

%% file: sections/section_conclusion.tex
\section{Conclusion}
\label{sec:conclusion}

In this report, we have introduced the Darwin Mobile Agent, an open-source infrastructure designed as a foundation for self-evolving agents in the mobile GUI domain. By treating the mobile ecosystem as a ``Big World'' for agent development, we move beyond the limitations of static, human-annotated datasets toward a framework that learns through autonomous interaction. Our work addresses the data-collection bottleneck in real-world mobile environments by providing a stable, asynchronous pipeline powered by cloud phones.

Alongside this infrastructure, we propose a conceptual roadmap for systematically removing human priors from three fundamental pillars of self-evolution: task curriculum, outcome verification, and agent state. Our initial results suggest that this framework provides the stability and scalability required for policy optimisation in complex, non-stationary environments, even under hardware instability. While the current implementation represents only the first stage of our proposed roadmap, it establishes the practical and theoretical foundation necessary for the eventual goal of the Darwin system: agents that autonomously evolve their capabilities through their own experience.

%% file: sections/appendix.tex
\clearpage
\section{Implementation Details}
\label{appendix:implement_details}

Across all experiments, we follow the online interaction--update loop described in \cref{subsec:rollout_loop}. Each run is executed on a multi-GPU server and typically takes 24 to 36 hours to complete. The hyperparameters are listed in \cref{tab:appendix_hparams}. The tasks for each experiment are selected from SPA-Bench \citep{chenspa}, which covers diverse everyday workflows (e.g. e-commerce, social, utilities) derived from production applications with visually dynamic interfaces and naturally occurring UI variations. It includes both English and Chinese instructions, enabling evaluation of multilingual agent capabilities. \cref{tab:spabench_tasks} reports the task subsets used in our experiments.

\begin{table}[h]
\centering
\caption{Training hyperparameters in the training runs.}
\label{tab:appendix_hparams}
\begin{tabular}{l c}
\toprule
\textbf{Hyperparameter} & \textbf{Value} \\
\midrule
\multicolumn{2}{l}{\textit{Data / sequence lengths}} \\
Max prompt length & 8192 \\
Max response length & 512 \\
Actor memory length (images) & 1 \\
\midrule
\multicolumn{2}{l}{\textit{Environment / rollout}} \\
Episode horizon $H$ (max steps) & 32 \\
Rollout steps per update & 256 \\
Action execution wait (seconds) & 2 \\
\midrule
\multicolumn{2}{l}{\textit{Optimisation (PPO)}} \\
Discount factor $\gamma$ & 1.0 \\
GAE $\lambda$ & 0.95 \\
Actor learning rate & $1\times 10^{-6}$ \\
Critic learning rate & $1\times 10^{-5}$ \\
Actor PPO epochs per update & 4 \\
Actor PPO mini-batch size & 64 \\
Micro-batch size per GPU & 16 \\
PPO clip ratio (high) & 0.2 \\
\midrule
\multicolumn{2}{l}{\textit{Critic warmup / stabilisation}} \\
Critic warmup epochs & 30 \\
Warmup PPO epochs per step & 4 \\
Warmup GAE $\lambda$ & 1.0 \\
\bottomrule
\end{tabular}
\end{table}

\begin{table*}[t]
\centering
\caption{Selected Spa-Bench tasks used in the experiments. The 16-task setting extends the base 8-task subset.}
\label{tab:task_list_spabench}
\scalebox{0.75}{
\begin{tabular}{l l}
\toprule
\textbf{App} & \textbf{Task} \\
\midrule
\multicolumn{2}{l}{\textbf{Base 8 Tasks}} \\
\texttt{Amazon} & \texttt{Get the search results for 'sunglasses'.} \\
\texttt{Booking.com} & \texttt{Get the search results for stays in Berlin. Select any date, rooms and guests.} \\
\texttt{Booking.com} & \texttt{Navigate to app settings.} \\
\texttt{Chrome} & \texttt{Clear all active tabs.} \\
\texttt{Clock} & \texttt{Set an alarm for 9am on weekdays.} \\
\texttt{Clock} & \texttt{Set Home time zone to 'Hong Kong'.} \\
\texttt{Settings} & \texttt{Go to notification settings. Turn on Notification History.} \\
\texttt{Settings} & \texttt{Go to display settings. Turn on Dark Theme.} \\

\midrule
\multicolumn{2}{l}{\textbf{Additional 8 Tasks added for the 16-task setting}} \\
\texttt{Chrome} & \texttt{Get the search results for Taylor Swift.} \\
\texttt{Clock} & \texttt{Add current time at London (UK) to clock.} \\
\texttt{Dictionary Merriam Webster} & \texttt{Look up the definition of the word 'agent'.} \\
\texttt{Google Play} & \texttt{Get the search results for WhatsApp.} \\
\texttt{Google Play} & \texttt{Check the details of General settings.} \\
\texttt{Settings} & \texttt{Go to display settings. Turn on Dark Theme.} \\
\texttt{X} & \texttt{Draft a post with the content 'Written by Agent1'.} \\
\texttt{X} & \texttt{Search for the account @Mayday\_EN. Follow it.} \\

\bottomrule
\end{tabular}
}
\label{tab:spabench_tasks}
\end{table*}

\section{LLM Judge Configuration and Performance}
\label{sec:judge_configs}

To identify a reliable reward signal for the reinforcement learning pipeline, we evaluated the alignment of the LLM judge against human-labelled ground truth. We conducted a feature ablation study using 296 task trajectories (147 English; 149 Chinese) to determine which prompt configurations best inform the verification process. All evaluations were performed using \texttt{gemini-2.5-flash}.

As shown in \cref{tab:llm_judge_results}, providing the complete interaction history is the most significant factor in judge accuracy. The baseline configuration, which includes the full trajectory of screenshots and actions, achieves an F1 score of 0.87 on English tasks and 0.82 on Chinese tasks, significantly outperforming the configuration that only evaluates the final state. We further evaluated three primary augmentations to this baseline: marked interaction (visual indicators of action coordinates), purpose analysis (inference of agent intent), and execution summaries (state change analysis).

For English-language tasks, adding only marked interaction coordinates yielded the highest alignment with human labels, achieving an F1 score of 0.91. For Chinese-language tasks, performance was highest when combining marked coordinates with execution summaries, yielding an F1 score of 0.85. Notably, adding multiple layers of explicit reasoning (e.g. purpose analysis combined with summaries) did not improve performance.

\begin{table*}[t]
\centering
\caption{Performance metrics for the LLM judge across all prompt and visual configurations.}
\label{tab:llm_judge_results}
\scalebox{0.82}{
\begin{tabular}{l c c c c c c c c}
\toprule
& \multicolumn{4}{c}{\textbf{English Tasks}} & \multicolumn{4}{c}{\textbf{Chinese Tasks}} \\
\cmidrule(lr){2-5} \cmidrule(lr){6-9}
\textbf{Configuration} & \textbf{F1} & \textbf{Acc} & \textbf{Prec} & \textbf{Rec} & \textbf{F1} & \textbf{Acc} & \textbf{Prec} & \textbf{Rec} \\
\midrule
Final Trajectory State Only & 0.78 & 0.76 & 0.86 & 0.71 & 0.70 & 0.80 & 0.76 & 0.65 \\
\textbf{Baseline (Full Trajectory)} & 0.87 & 0.85 & 0.90 & 0.84 & 0.82 & 0.87 & 0.86 & 0.78 \\
\midrule
+ Marked Only & \textbf{0.91} & 0.90 & 0.93 & 0.90 & 0.83 & 0.89 & 0.89 & 0.78 \\
+ Purpose Only & 0.88 & 0.86 & 0.88 & 0.88 & 0.80 & 0.87 & 0.85 & 0.76 \\
+ Summary Only & 0.90 & 0.88 & 0.92 & 0.88 & 0.82 & 0.88 & 0.91 & 0.74 \\
\midrule
+ Marked + Purpose & 0.88 & 0.86 & 0.93 & 0.83 & 0.76 & 0.84 & 0.84 & 0.69 \\
+ Marked + Summary & 0.87 & 0.86 & 0.91 & 0.84 & \textbf{0.85} & 0.89 & 0.88 & 0.81 \\
+ Purpose + Summary & 0.88 & 0.86 & 0.91 & 0.85 & 0.78 & 0.86 & 0.88 & 0.70 \\
+ All Features & 0.89 & 0.88 & 0.90 & 0.88 & 0.80 & 0.87 & 0.93 & 0.70 \\
\bottomrule
\end{tabular}
}
\end{table*}

\section{Action Space}
\label{appendix:action_space}
We define a discrete action space that captures the fundamental interactions an agent can perform on a mobile device. Each action is parameterised by explicit arguments (e.g. coordinates, directions, or text content), enabling precise, executable control. The complete set of supported actions and their corresponding parameters is summarised in Table~\ref{tab:action_space}.

\begin{table}[h]
\centering
\begin{tabular}{ll}
\toprule
\textbf{Action} & \textbf{Definition} \\
\midrule
\texttt{click(x, y)} 
& Clicks at screen coordinates $(x, y)$. \\

\texttt{long\_press(x, y)} 
& Performs a long press at coordinates $(x, y)$. \\

\texttt{type(content)} 
& Types the specified text content. \\

\texttt{scroll(x, y, direction)} 
& Scrolls the screen at $(x, y)$ in the given direction. \\

\texttt{drag(x$_1$, y$_1$, x$_2$, y$_2$)} 
& Drags from $(x_1, y_1)$ to $(x_2, y_2)$. \\

\texttt{open\_app(app)} 
& Launches the specified application. \\

\texttt{press\_home()} 
& Presses the home button. \\

\texttt{press\_back()} 
& Presses the back button. \\

\texttt{wait()} 
& Pauses execution to allow the UI state to update. \\

\texttt{finished()} 
& Marks the task as complete. \\
\bottomrule
\end{tabular}
\caption{Action space of the agent and their corresponding definitions.}
\label{tab:action_space}
\end{table}